\pdfoutput=1
\documentclass[11pt]{article}

\usepackage[final]{acl}

\usepackage{times}
\usepackage{latexsym}

\usepackage[T1]{fontenc}

\usepackage[utf8]{inputenc}

\usepackage{microtype}

\usepackage{inconsolata}

\usepackage{graphicx}
\usepackage{amsmath}
\usepackage[symbol]{footmisc}
\usepackage{csquotes}
\usepackage{epigraph}
\usepackage{listings}
\usepackage{hyperref}
\usepackage[greek,english]{babel}
\usepackage{textgreek}
\usepackage{booktabs}

\usepackage[most]{tcolorbox}
\usetikzlibrary{shadows.blur}

\usepackage{tikz}
\usetikzlibrary{positioning, arrows.meta, shapes, positioning}

\usepackage{enumitem}

\usepackage{amssymb}
\usepackage{pifont}
\tcbset{
    mymsg/.style={
        enhanced,
        colframe=white, frame hidden,
        boxrule=0pt,
        arc=18pt,
        outer arc=18pt,
        boxsep=2pt, left=8pt, right=8pt, top=6pt,
        colback=#1!25,
        fuzzy shadow={1mm}{-0.7mm}{0mm}{0.1mm}{black!40!white},
        width=0.8\linewidth,
        before skip=-0.5mm, 
    },
    expertstyle/.style={mymsg=green, enlarge left by=10mm},
    llmstyle/.style={mymsg=blue, enlarge right by=10mm}
}

\title{Investigating Expert-in-the-Loop LLM Discourse Patterns for Ancient Intertextual Analysis}

\author{Ray Umphrey\footnote[2]\\ \\
  University of the Cumberlands \\
  \texttt{ray.umphrey@ucumberlands.edu} 
  \\\And
  Jesse Roberts\footnote[2]\\ \\
  Tennessee Tech University \\
  \texttt{jtroberts@tntech.edu} 
  \\ \AND
  Lindsey Roberts \\
  Tennessee Tech University \\
}

\setlist{nosep}

\begin{document}
\maketitle
\begin{abstract}

This study explores the potential of large language models (LLMs) for identifying and examining intertextual relationships within biblical, Koine Greek texts. By evaluating the performance of LLMs on various intertextuality scenarios the study demonstrates that these models can detect direct quotations, allusions, and echoes between texts. The LLM's ability to generate novel intertextual observations and connections highlights its potential to uncover new insights. However, the model also struggles with long query passages and the inclusion of false intertextual dependences, emphasizing the importance of expert evaluation. The expert-in-the-loop methodology presented offers a scalable approach for intertextual research into the complex web of intertextuality within and beyond the biblical corpus.

\end{abstract}

\section{Introduction}
    \footnotetext[2]{Equal Contribution}
    

Intertextuality, coined by Julia Kristeva \cite{kristeva1980desire}, proposes that the meaning which should be understood as intended by an author is that most relevant in the common zeitgeist. However, like stacked layers of a fuzzy neural network \cite{kwan1994fuzzy}, authors' work transforms the meaning of the input linguistic ``sign'' so that the understanding incumbent on future works is forever changed.

As shown in Figure \ref{fig:intertextuality_img}, many contemporary references to Satan depend more (but not exclusively) upon Milton than the Bible, while Genesis provides the base understanding invoked and transformed by Milton \cite{allen2011intertextuality}. This notion of an intertextual network is an extension of the intertextual graphs discussed in \citet{kuznetsov2022revise} as it supposes a non-linear transformation is applied by the author in the work to the inherited linguistic sign.

\begin{figure}[h]
    \centering
    \resizebox{\linewidth}{!}{
    \begin{tikzpicture}[
            auto,
            font=\sffamily,
            every node/.style={
                draw,
                fill=white,
                text centered,
                minimum height=0.5em,
                font=\sffamily\small,
                inner sep=0.5em,
                rounded corners=5pt
            },
            arrow/.style={
                -{Latex[length=2mm, width=1.5mm]},
                line width=1.5pt,
                color=black,
            },
            node content/.style={
                align=center
            },
            label style/.style={
                font=\sffamily\small,
                inner sep=1pt,
                draw=none
            }
            ]
    
        \node[node content] (genesis) {Genesis};
        \node[node content, above right=0.5cm of genesis] (paradise) {$f_1(\sum_{\forall i}(x_i))=$ Paradise Lost };
        \node[node content, below right=1cm of genesis] (hocus) {$f_2(\sum_{\forall j}(x_j))=$ Hocus Pocus};
        
        \draw[arrow] (genesis) -- (paradise) node[midway, left, label style] {1};
        \draw[arrow] (paradise) -- (hocus) node[midway, right, label style] {0.9};
        \draw[arrow] (genesis) -- node[midway, left, label style] {0.1} (hocus);

    \end{tikzpicture}
    }
    \caption{Example approximated intertextual dependence over Genesis, Paradise Lost \cite{milton2005paradise}, and Hocus Pocus \cite{HocusPocus1993}. Future works depend on past related texts with some weight. Subsequent work transforms the combined inherited weighted representations like a neural activation function.}
    \label{fig:intertextuality_img}
    \vskip-1em
\end{figure}

When the date of authorship and lineage of a text are known, the structure of the network can be asserted and the weights may be studied to understand intertextualities. However, in biblical studies, texts are often dated imprecisely or may not have been sufficiently distributed to support confident assertions of textual dependence. Therefore, the presence of a strong intertextual dependence can provide both important situational and interpretational guidance for the passage.

While many such intertextual depencies have been documented, the relative youth of the interpretational framework and the vast number of relatively co-temporal texts makes searching for potential intertextualities tedious.

\begin{table}[h]
    \caption{List of Contributions}
    \label{tab:contributions}
    \resizebox{\linewidth}{!}{
    \begin{tabular}{c p{6.5cm}} 
    
     \toprule
    \textbf{Contribution 1} &
        \small \textit{We find that the query length has a significant impact while long corpora pose no issue to retrieval.} \\ 
        \hline
    \textbf{Contribution 2} & 
        \small \textit{We identify a pattern of LLM usage that augments an expert's ability to catalogue and evaluate the weight of intertextual dependencies in Koine Greek.} \\ 
        \hline
    \textbf{Contribution 3} &
        \small \textit{We identify types of errors made by the LLM.} \\ 
        \hline
    \textbf{Contribution 4} &
        \small \textit{We show that LLMs perform intertextual analysis by retrieving known intertextualities and by directly comparing the similarity of passages.} \\ 
        \hline
    \textbf{Contribution 5} &
        \small \textit{We identify a strong previously undocumented intertextuality through LLM support.} \\ 
        \bottomrule
    \end{tabular}
    } 
\vskip-0.5em
\end{table}

This paper seeks to use large language models (LLMs) to identify and examine the strength of intertextual relationships within biblical texts. By using LLMs to detect intertextuality in biblical writings, the user can establish patterns of usage by individual biblical authors and shed light on how texts were understood and reinterpreted by later writers and illuminate the understanding transmitted by the author from inherited concepts. 

Armed with an appropriate pattern of LLM usage, researchers may be able to scale their efforts and shed light on long-standing questions in biblical studies. Further, the methodology can be applied to texts and contexts outside the biblical corpus. 


In the following sections we provide the necessary background regarding LLMs and intertextual analysis, discuss the experiments used to substantiate the contributions in Table \ref{tab:contributions}, and describe the experimental results. We conclude with a discussion of the limitations of this work.

\section{Background}


Richard Hays’ seminal work, \textit{Echoes of Scripture in the Letters of Paul}, established intertextuality as a central concern of biblical interpretation for subsequent research \cite{hays1989echoes}. The interest in biblical intertextuality has only grown through the years, yet a lack of standardized definitions and methodology remains \cite{emadi2015intertextuality}. 

Discerning intertextual relationships in biblical texts is not always a straightforward endeavor. Biblical texts often reference previous writings, both biblical and extra-biblical. Biblical authors are frequently creative in how they use other texts and do not always cite their sources. Often biblical authors will merely allude to, paraphrase, or creatively embed intertextual elements in their writing. This makes intertextual studies challenging and, at times, controversial.  

The study of biblical intertextuality is relevant to understanding the relationship of the Old Testament to the New Testament, how later biblical writers understood and interpreted previous texts, and how much contemporary writers knew of other current writings. One prominent area of intertextual study is the synoptic problem, which addresses the relationships of the three synoptic gospels to one another in relation to their interdependence, sources, and manner of composition. Intertextuality has also paved the way for fresh analyses of the relationship between Paul and Jesus \cite{wenham1995paul}. Advances in intertextual analysis through AI and LLMs have the potential to illuminate such important issues in biblical studies. 

\subsection{LLMs}

Transformer-based large language models (LLMs) like Claude \cite{anthropic2024claude} have been shown to be remarkably capable of language analysis. LLMs function by generating context aware embeddings of the tokenized words in the context and generating an attention-filtered compression given the current last word in the sequence. The compression is then fed to a non-linear function in the form of a neural network. This process of projection, attention, compression, and non-linear mapping is repeated serially through the layers of the LLM. 

Projection into an embedding space has been shown to permit semantic and syntactic reasoning \cite{mikolov2013efficient} through a continuous vector representation. Further, attention has been shown to enable more long range dependencies and meanings to be captured and conveyed into the compressed representation of input text \cite{bahdanau2014neural}. Finally, the feed-forward non-linear mapping provides more efficient learning from the available data \cite{vaswani2017attention} and universal computation over the input text \cite{roberts2023powerful}. 

While the technological innovation associated with transformers \cite{vaswani2017attention} is arguably an increment over natural language processing (NLP) techniques that introduced word vectors \cite{mikolov2013efficient} and attention \cite{bahdanau2014neural}, LLMs have proved to be a revolutionary advance in the performance of virtually all NLP tasks. 

\subsubsection{Related Work}

In \citet{coffee2012tesserae}, the authors specifically sought to detect textual allusions by identifying shared words between two texts. They found this method was able to identify previously uncatalogued passages that may contain allusions. While this work was an important step, it is brittle to superficial word overlap and does not have the ability to reason about more nuanced contact between the texts (themes, synonyms, etc).

\citet{daietal2023llm} and \citet{yu2024assessing}  used GPT-3.5 and GPT-4 \cite{radford2019language} to do thematic coding. In each, the authors compare the theme label generated by the LLM to the label generated by a human annotators. They show that the tested LLMs tend to be able to reason over thematic content. Finally, \citet{khan2024automating} performed LLM assisted corpus coding for function-to-form pragmatic and discourse analysis.

The existing work demonstrates compelling progress toward scalable, automated reasoning for qualitative textual analysis with human collaboration. However, our work importantly augments the existing literature in four ways: 1) direct qualitative intertextual comparison by an LLM, 2) the task requires retrieval rather than labeling, 3) the prompt pattern is intended to augment rather than automate human ability, and 4) texts are presented in a non-English language (Koine Greek).

\section{Definitions and Criteria}

This paper is concerned with identifying and retrieving specific literary dependencies between biblical texts in Koine Greek using Claude Opus \cite{anthropic2024claude} which can be evaluated by the expert-in-the-loop to understand intertextual dependence and weight. Due to the multivalent nature of biblical intertextuality, the dependencies may appear as direct quotations, allusions, or echoes \cite{hays1989echoes}. Allusion can be defined as an indirect reference with some definable characteristics, such as lexical or thematic correspondence. Echoes, however, are subtle references that may exist purely on a structural or conceptual level or a single specific keyword.  While direct quotations are usually apparent to the reader, allusions and echoes require some criteria to help with identification. 

\citet{hays1989echoes} enumerates seven criteria for determining allusions and echoes: availability of the source to the author, volume (characterized by repetition, distinct patterns, prominence), recurrence of the citation by the author, thematic coherence, historical plausibility, history of interpretation, and satisfaction or sensibility. For this study, we will use Hays’ criteria as the expert lens for human evaluation of candidate intertextualities identified by the LLM.

\section{Methodology}

We identify 6 passages each representing a unique scenario of interest. We provide them to Claude Opus via the pattern in Figure \ref{fig:LLM_interaction} and receive responses similar to that shown. The returned candidate intertextualities are then evaluated for strength based on the criteria discussed above. 

In support of the contributions listed in Table \ref{tab:contributions}: We 1) evaluate the effect of query and corpus length by varying each while maintaining the passage and prompt pattern. We 2) evaluate all passages in Koine Greek with an author specialized to intertextual biblical analysis as the expert-in-the-loop providing evaluation over the candidates in the form of a fuzzy weight (labeled as low, mid, or high) as alluded to in Figure \ref{fig:intertextuality_img}. We 3,4, and 5) identify failure and performance modes across a comprehensive set of scenarios that present passages with expected false positive, known true positive, probable, speculative, hypothetical, and out-of-sample true positive intertextualities.

\begin{figure}[h!]
    \centering
    \resizebox{\linewidth}{!}{ 
    \begin{tcolorbox}[expertstyle]
        \hfill \textbf{Expert} \\
        I am looking for intertextualities, that is quotations, allusions, or general dependence on another passage, between two passages. Identify rare words that are mutually used as well. Please denote potential intertextualities between the corpus and query passages below: \\
        \textcolor{black}{[Corpus]} \\
        \textcolor{black}{[Query]}
    \end{tcolorbox}
    } 

    \vspace{2mm} 

    \resizebox{\linewidth}{!}{ 
    \begin{tcolorbox}[llmstyle]
        \textbf{LLM} \\
        Here are the potential intertextualities I identified between the two passages:\\
        1.	The phrase “ἐν ᾧ μέτρῳ μετρεῖτε μετρηθήσεται ὑμῖν” in the corpus passage is very similar to “ᾧ μέτρῳ μετρεῖτε, ἐν αὐτῷ μετρηθήσεται ὑμῖν” in the query. This appears to be a direct quotation.\\
        2.	Both passages use the relatively uncommon future passive verbs “κριθήσεσθε” (you will be judged) and “μετρηθήσεται” (it will be measured). ... \\
        5.	Both exhort the audience to act/treat others in a certain way, using the 2nd person plural present active verbs π\textomikronιεῖτε (you do/make) and δίδ\textomikronτε (you give).\\

    \end{tcolorbox}
    } 
    \caption{Prompt pattern used to identify intertextualities between the corpus (to be searched) and the query (passage of interest).}
    \label{fig:LLM_interaction}
\vskip-0.5em

\end{figure}

\subsection{Selected Texts and Scenarios}

\begin{table}[h]
\centering
\caption{Test Scenarios and corresponding corpus and query passages.}
\resizebox{\linewidth}{!}{%
\begin{tabular}{l c c}
\toprule
Test Scenario & Corpus & Query \\
\hline
True Positive & Matthew 7 & 1 Clement 13 \\
False Positive & Matthew 7 & 1 Peter 2:4-8 \\
Probable & Sirach 51 & Matthew 11:25-30 \\
Speculative & Romans 3:19-5:11 & Luke 18:9-14 \\
Hypothesis 1 & Matthew 10:5-42 & Acts 20:17-35 \\
Hypothesis 2 & Matthew 10:5-42 & Luke 10:2-12 \\
Out-of-Sample & Matthew 7 & Fake Biblical Stylized Text\\
\bottomrule
\end{tabular}%
}
\label{tab:test_rationales}
\vskip-0.5em
\end{table}

Matthew 7 is tested against two texts, a known positive and a known passage with superficial similarity, to evaluate the model’s sensitivity in the cases of true positive and expected false positive. In 1 Clement 13, the author directly quotes from Matthew 7:1-2 and attributes the quote to Jesus. 1 Peter 2:4-8 contains no known parallel with Matthew 7 but shares a common stone metaphor with Matthew 7:24-27.

The remaining four sets are all experimental or hypothetical in some way. First, similar to the true positive, some scholars argue that Matthew 11:25-30 depends on Sirach 51 while others merely acknowledge the similarities \cite{Hagner_1993}. On the other hand, most scholars reject dependence between Romans and Luke 18:9-14 but acknowledge superficial thematic similarity \cite{johnson1991gospel}. 

We suspect Matthew 10:5-42 to have literary interdependence with Acts 20:17-35 based on novel, unpublished research. However, no known proponent of this relationship exists. Interestingly, the implications of this scenario can potentially inform the debate surrounding the synoptic problem. 

Finally, to test if LLMs can identify truly novel intertextualities apart from potential knowledge from an unknown pretraining corpus, we develop a novel passage that shares features with Matthew 7 regarding a fish and a tree. 

While intertextuality can exist across texts composed in different languages, we chose texts composed in Koine Greek to keep the project manageable. Since intertextuality often depends on lexical forms, unique vocabulary, and morphological features, we conducted the exercises in Greek rather than English translations. Further, research in the field of new testament studies is conducted in Greek , and we hope LLMs will be used to augment the work of other researchers in this space.

\section{Experiments}

\subsection{True Positive: Matthew 7 \& 1 Clement 13}

\textbf{Background:} In 1 Clement 13, the author seems to synthesize several related sayings that can be found throughout the Sermon on the Mount \cite{holmes2007apostolic}. Scholars date the writing of 1 Clement as being after or contemporary with the composition of Matthew’s gospel. While it is possible that the authors of both texts draw from a hypothetical common source, no such document exists, and the grouping of the same sayings in both texts suggests some direct dependence \cite{holmes2007apostolic}. 

\textbf{Candidate Intertextualities:} The language model identified the phrase, ``With the measure you use, it will be measured to you,'' as a direct quote (Matt. 7:2). In Greek, the only variation from the Matthean text is the omission of the preposition ἐν (``by the measure you use'') and the addition of the prepositional phrase ἐν αὐτῷ (``by this it will be measured''). The model detected other lexical parallels, such as the use of the verb κρίνετε ``you judge'' and κριθήσεσθε ``you will be judged,'' as well as the usage of δίδωμι ``give'' (Matt. 7:7). 

The model also detected morphological similarities in verbs used, such as the use of the future passive forms μετρηθήσεται ``it will be measured'' and κριθήσεσθε ``you will be judged.'' It also detected identical imperative forms in person, number, tense, and voice: π\textomikronιεῖτε (you do) and δίδ\textomikronτε (you give). The model detected general thematic and structural features common to both texts. 

\textbf{Observations \& Analysis:} The 1 Clement citation explicitly credits Jesus with the sayings (``Let us remember the words of the Lord Jesus''). Interestingly, the model mentioned sayings not included in Matthew chapter 7 but did not locate them. This indicates that the model was working within the boundaries of the provided texts but utilizing broader knowledge acquired during pre-training. When we reran the exercise and omitted the explicit reference to Jesus, the model did not acknowledge the additional sayings. The language model generated no false positives and the weight of the intertextual dependence is high.

\subsection{False Positive: Matthew 7 \& 1 Peter 2:4-8}

\textbf{Background:} 1 Peter bears no evidence of direct literary dependence on the synoptic gospels. However, both Matthew 7:24-27 and 1 Peter 2:4-8 use stone metaphors. This provides an opportunity to identify model sensitivity to false positives. As expected, the model retrieved the superficial textual similarities. 

\textbf{Candidate Intertextualities:} The model concluded that 1 Peter 2:4-8 drew on the imagery and language of Matthew 7. It cited the use of construction metaphors, the use of πέτρα ``rock/stone'' and πρ\textomikronσκόπτω and πρόσκ\textomikronμμα ``stumble/stumbling stone,'' and the theme of ``two ways'' in both passages. However, the model’s assertion of direct dependence can be confirmed as a false positive since the pertinent references in 1 Peter 2:4-8 are explicit quotations from the Old Testament \cite{marcar2016quotations}. 


\textbf{Observations \& Analysis:} While the model provided a false positive, the exercise was helpful. First, the model built a convincing case with several concrete and sound data points. There are striking similarities between Matthew 7 and 1 Peter 2:4-8. Additionally, the model identified the motif of a rejected stone in both passages. This is the explicit content of the Psalm 118 citation in 1 Peter 2:7, but the model detected the same motif in Matthew 7:9. 

However, the shared Old Testament tradition best accounts for these similarities since the stone motif from Isaiah and Psalms is extensively cited in the New Testament. This gives warrant to further research into the potential intertextual link between the Isaiah and Psalms texts with Matthew 7. The language model generated valid candidates with no false positives. The weight of the intertextual dependence is low, but likelihood of shared dependence with a common pretext is mid to high. 


\subsection{Probable: Sirach 51 \& Matthew 11:25-30}

\textbf{Background:} The Wisdom of Ben-Sirach, or Sirach, is a second-century B.C.E. Jewish wisdom book in the Deutero-canonical (or apocryphal) writings. Some scholars maintain that Sirach 51 provides the background to Jesus’ words in Matthew 11:25-30, but it is not a universally affirmed intertextual connection \cite{Hagner_1993}. This experiment will confirm whether LLMs can contribute to a positive case for intertextuality between these writings. 

\textbf{Candidate Intertextualities:} The model listed several anticipated intertextual links, such as the theme of revealed wisdom, references to the ``yoke'' of wisdom, instruction, and finding rest. These were valid intertextual candidates and are supported by prominent commentators \cite{luz2001matthew}. The most meaningful contribution was related to the structure of the passages. The model noted that both passages begin with thanksgiving and an acknowledgment of God as father (Sirach 51:1, 10; Matthew 11:25-27). It further elaborated that the rhetorical form consisting of a prayer to God followed by an exhortation to seek wisdom was an established pattern in Jewish wisdom literature. The LLM neither confirmed nor denied a direct intertextual relationship between the corpus and query but suggested that the passages draw on similar rhetorical patterns.

\textbf{Observations \& Analysis:} Our interest turned to the rhetorical form of the passages and where else this form might occur. When prompted, the LLM provided examples of biblical and extra-biblical texts that reflect this pattern and cited relevant existing research. In this way, the model generated candidate texts for further study. We are unaware of anyone making this distinction regarding the relationship between Sirach 51 and Matthew 11:25-30. Also, there appears to be no established nomenclature for this rhetorical phenomenon. 

Further research could shed light not only on the question of intertextuality but also on the literary unity of Sirach 51. This chapter is composed of three poems, and the question of its unity and history of composition is unsettled \cite{goodman2012apocrypha}. The insights of the language model have revealed a connection between issues of intertextuality, Jewish wisdom rhetoric, and literary composition, which merits further research.

The language model generated valid candidates with no false positives and the weight of the intertextual dependence is mid.


\begin{table*}[t]
\centering
\resizebox{\linewidth}{!}{%
\begin{tabular}{p{0.3\textwidth} c p{0.31\textwidth} c p{0.3\textwidth}}
\toprule
Acts 20:17-35 & Connection type & Matthew 10:5-42 & Connection type & Luke 10:2-12 \\
\toprule\addlinespace
19, 23 Paul's recounts his suffering & Thematic & 17-19, 23 Jesus promises suffering & & \\
\hline\addlinespace
22-23 Holy Spirit/persecution & Lexical \& Thematic & 20 Holy Spirit/persecution & & \\
\hline\addlinespace
23 in every city & Lexical \& Thematic & 5, 11, 14-15 whatever city you enter & Lexical \& Thematic & 8, 10, 12 whenever you enter a city \\
\hline\addlinespace
24 Paul counts his life of no value & Lexical \& Thematic & 39 Whoever loses his life finds it & & \\
\hline\addlinespace
24 testify & Lexical & 18 testify & & \\
\hline\addlinespace
25 proclaiming the kingdom & Lexical & 7 proclaim… the kingdom… & Lexical & 9 the kingdom has come \\
\hline\addlinespace
28-29 church as flock & Thematic & 6 Israel as sheep & & \\
\hline\addlinespace
29 wolves among sheep & Lexical \& Thematic & 16 sheep among wolves & Lexical \& Thematic & 3 sheep among wolves \\
\hline\addlinespace
33 silver, gold, apparel & Lexical \& Thematic & 9-10 gold, silver… two tunics & Thematic & 4 moneybag, knapsack, sandals \\
\hline\addlinespace
35 Paul's hard work & Thematic & 10 worker worthy of food & Thematic & 7 worker worthy of wages \\
\hline\addlinespace
35 help the weak ἀσθεν\textomikronύντων & Lexical & 8 heal the sick ἀσθεν\textomikronῦντας & Lexical & 9 heal the sick ἀσθεν\textomikronῦντας \\
\hline\addlinespace
35 more blessed to give than to receive & Lexical & 8 freely you have received, freely give & & \\
\bottomrule
\end{tabular}%
}
\caption{Candidate Intertextualities for Acts 20:17-35 to Matthew 10:5-42 and Luke 10:2-12 to Matthew 10:5-42}
\label{tab:passage_comparison}
\vskip-0.5em
\end{table*}

\subsection{Speculative: Romans 3:19-5:11 \& Luke 18:9-14}

\textbf{Background:} This experiment involves the relationship of Jesus' sayings to the teachings of Paul. Intertextual analysis of Jesus' sayings and Paul's writings has produced fruitful results, leading \citet{wenham1995paul}, \citet{allison1982pauline}, and others to posit a close connection between Jesus and Paul. The parable in Luke 18:9-14 tells us that the sinful tax collector was ``justified'' while the Pharisee was not. Here, Luke uses the verb δικαιόω ``justify'' in a similar manner as Paul when he writes of justification by faith in Romans. The scholarly consensus is that while there is an overlap of ideas, there is no literary interdependence between these two texts \cite{johnson1991gospel}. We chose these texts to learn whether LLMs could provide evidence of intertextuality.  

The corpus passage, Romans 3:19-5:11, was chosen based on the distribution of the word δικαιόω in Romans. The word occurs 15 times in Romans; the selected portion contains nine occurrences. When we asked the model to narrow the corpus passage to the most concentrated section of potential intertextual connections, it identified Romans 3:21-4:8 as the portion with the highest density and differentiated it from the subsequent paragraph about Abraham. Seven of the nine occurrences of δικαιόω in the original corpus occur within the narrowed corpus. By narrowing the corpus, the LLM conducted what \citet{guthrie1993structure} calls "cohesion shift analysis" which detects shifts in "cohesion fields" usually around paragraph breaks.

\textbf{Candidate Intertextualities:} The LLM picked up on the repeated key terms δικαιόω ``justify'' and ἁμαρτωλός ``sinner'' as well as their cognate noun forms δικαι\textomikronσύνη ``righteousness/justice'' and ἁμαρτία ``sin'' which occur frequently in the corpus. The LLM made loose thematic connections as well, such as the central role faith plays in both passages (although the Greek word πίστις is not used in Luke 18:9-14). The model rightly noted that both passages feature righteousness apart from works or boasting. The Pharisee in Luke 18:9-14 illustrates the concept which is explicit in Romans 3:21-4:8.

The LLM rightly notes the occurrence of a rare word in both passages. In Romans 3:25, Christ is a ἱλαστήρι\textomikronν ``propitiation'' for sin, and in Luke 18:13, the tax collector asks God to ἱλάσθητί ``propitiate'' him, the sinner (ἁμαρτωλός). The words are cognates: a noun in Romans and a verb in Luke. Both words only occur twice in the New Testament (the other occurrences are in the Epistle to the Hebrews). The LLM curiously cited references to Abraham as an intertextual candidate, while there is no reference to Abraham in Luke 18:9-14.

We then asked the LLM to locate potential intertextual references to Romans 3:21-4:8 in the preceding and subsequent pericopes. Using Luke 18:1-8 as the new query, the LLM found two lexical correspondences. The use of the word πίστις ``faith'' in Luke 18:8 and two cognate words with δικαιόω/δικαι\textomikronσύνη, ἐκδίκησόν meaning ``grant justice'' and ἀδικίας which describes the ``unrighteous'' judge.

Next, we asked the LLM to analyze Luke 18:15-27, which follows the original query passage. It provided three meaningful connections. It juxtaposed the emphasis on righteousness apart from the law in Romans to the focus on keeping the law in Luke 18:20-21. The LLM also noted the correspondence between the righteousness of God in Romans and the statement in Luke 18:19 that ``none is good except God alone.'' The LLM correlated the statement in Luke 18:27 that nothing is impossible with God with the claim in Romans 4:5 that God justifies the ungodly, which is impossible with man. 

\textbf{Observations \& Analysis:} 

While this exercise did not yield groundbreaking insights, the LLM provided valuable analysis of the passages in question and handled multiple queries well. When asked to evaluate the three queries in terms of their relatedness to the corpus, the model identified the original query as the most related. However, at no point did the LLM assert direct literary dependence between the corpus and queries.

The language model generated valid candidates with one false positive and the weight of the intertextual dependence is low to mid. 

\subsection{Hypothesis: Matthew 10:5-42, Acts 20:17-35, \& Luke 10:2-12}

\textbf{Background:} The rationale behind this exercise is exploratory. While working on another project, we found a journal article that noted similarities between Paul’s farewell discourse in Acts 20:17-35 and Jesus’ commissioning of the disciples \cite{brown1963synoptic}. Upon analyzing the passages, we suspected direct literary dependence between Matthew 10:5-42 and Acts 20:17-35 though no known scholarship explores intertextual connections between these passages. Since the passages in Matthew and Luke are parallel passages, we ran both against Acts 20:17-35 to see which had the strongest connections. Interestingly, Acts 20:17-35 and Luke 10:2-12 do not appear to have as strong of an intertextual dependence even though they share an author.

\textbf{Observations \& Analysis:} While the model acknowledged the possibility of literary dependence with both, it cited six intertextualities with Luke and twelve with Matthew. This is mainly due to Matthew’s expanded version of the missionary discourse. Matthew’s version contains unique material not included in Luke but alluded to in the Acts discourse. This is a remarkable observation considering that Luke and Acts have the same author, and Luke’s version of the missionary discourse omits much of the Matthean material present in Acts 20:17-35. 

When using Acts 20:17-35 as the query passage, the LLM provided several parallels and concluded that the passage was dependent on the Matthew passage. However, we continued submitting prompts and encountered false information and non-sequitur reasoning. We determined that this was due to the query size. We broke the query passage into smaller sections and ran each section separately. Reducing the query size also reduced the number of parallels, and the LLM did not recognize patterns of literary dependence when working through the query one section at a time. 

The language model generated over a dozen valid candidates with some false positives and the weight of the intertextual dependence is high. 

\subsection{Out-of-Sample: Matthew 7 \& Fish and the Tree}

\textbf{Background:} This comparison is intended to evaluate the LLM's ability to retrieve intertextualities when it is guaranteed that the model has no background information from pre-training on which to draw intertextual information. A novel, moralistic story stylized via an LLM was written in Koine Greek involving a tree and a fish. This story intentionally included motifs and words from the Matthean passage.

\textbf{Candidate Intertextualities:} The model generated the following intertextual candidates: the metaphor of a tree bearing fruit, the word ἰχθύς (fish), the thematic similarity between deep roots and a rock foundation, and the thematic similarity between deep roots of faith and good fruits of faith. The model generated two false positives by identifying the words π\textomikronταμός (river) and ἀγαθός (good) as rare words. The word ἀγαθός occurs 101 times in the New Testament while π\textomikronταμός occurs 17 times.

\textbf{Observations \& Analysis:} The model correctly identified intertextual similarities even though the query passage was entirely out of context. This shows with certainty that LLMs are capable of intertextual candidate retrieval without apriori knowledge. 

\textbf{English Translation of the Parable of the Fish \& the Tree:} The river flows through the forest, and by the river there is a great tree. But the tree bore good fruit and gave shade to the animals of the forest. But in the river there is a small fish, which asks for food every day. And the fish saw the tree and its fruits falling into the water. And he ate of the fruit and gave thanks to the tree. But there was a great flood and the river was flooded. But the river dragged the fish away from the tree. And the fish was troubled and afraid, having no food or shelter. But the tree had deep and strong roots, and it remained firm in the flood. And when the water receded, the fish found its way to the tree again. And they lived in peace, the tree providing and the fish giving thanks. Likewise we, if we have a deep root of faith, remain firm in tribulations and will find the way to God again. Because He always provides us with His goods according to the measure of our faith.

\section{Results and Observations}
The LLM successfully identified lexical correspondences by detecting common words and analyzing morphological data in verbal forms, including tense, mood, voice, person, and number. It also provided keyword statistics, sometimes identifying rare words and the number of occurrences in the New Testament. The model identified direct quotes, even when the quoted form was adapted or paraphrased. It made an important distinction between direct quotes and verbatim citations. Discerning a direct quote not in verbatim form shows advanced language processing capability. In addition to detecting these textual phenomena, the LLM also identified areas where these features were most dense within a larger corpus. 

In addition to lexical and textual analysis, the LLM found intertextual relationships through other modes of analysis. The model performed contextual analysis of the passages. It could differentiate between the usage of the same word in two different contexts. The same word used in a different context from the corpus text was not afforded the same weight as a word used in a similar or identical context. Additionally, the model could detect thematic correspondence when no lexical parallel was present. It also detected shared structural features that indicate relatedness between texts. 

It is worth noting that the level of analysis conducted by the LLM has great potential beyond intertextual study. The type of structural, lexical, and morphological analysis modeled here has been productively used for discourse analysis \cite{umphrey2022cohesive}. Like intertextual studies, discourse analysis suffers from a lack of clear methodological consensus and could benefit from the advancements offered by LLMs. 

\subsection{Evidence of Novel Intertextual Work}
One concern of this type of experiment is that the language model would draw from background knowledge in the training data rather than conducting novel analyses of the provided texts. On multiple occasions, the LLM demonstrated the ability to reason exclusively from provided texts yet draw from its knowledge bank as necessary. When analyzing the sayings of Jesus in 1 Clement 13, it acknowledged that some of the sayings were from outside the corpus text of Matthew 7 but did recognize the previous chapters of the Sermon on the Mount as their source (Matthew 5-7). This indicates that the LLM was doing a closed analysis of the provided texts. Similarly, the LLM asserted intertextual dependence between 1 Peter 2:4-8 and Matthew 7, when the correspondence points are direct quotations from Isaiah and Psalms. The model reasoned from the provided texts without looking outside them. This suggests the LLM tended to limit the scope of consideration, avoiding irrelevant retrieval beyond the target passage.

The analysis of Sirach 51 provided what appears to be a completely novel intertextual observation regarding the structure of both passages. Furthermore, the intertextual links discovered in the analysis of Acts and Matthew seem to be without precedent in the scholarly literature. These observations have potential for future research. We are convinced that LLMs have great potential for generating novel research ideas for biblical intertextual studies.

Finally, by retrieving intertextual candidates for an unseen passage, the LLM has shown that it is able to reason over complex intertextualities without the benefit of pre-trained knowledge.


\section{Conclusions}

This study demonstrates the potential for using LLMs to identify and examine intertextual relationships within biblical texts. By evaluating the effect of query and corpus length, testing performance on passages in Koine Greek, and assessing the model's ability to handle various intertextuality scenarios, we have shown that LLMs can be a valuable tool for biblical scholars. This tool works exceptionally well when the query is short and the corpus is between 1 and 3 chapters. By successive application of the pattern across corpora, this method can be used to evaluate large bodies of texts for intertextual connections to a query passage in a scalable manner.

Our findings suggest that LLMs are capable of detecting direct quotations, allusions, and echoes between biblical texts, even when presented in a non-English language. The LLM successfully identified lexical correspondences, morphological similarities, direct quotes (even when adapted or paraphrased), and thematic and structural parallels. It also demonstrated the ability to narrow down a corpus to the most relevant sections for intertextual analysis.

Importantly, the LLM exhibited evidence of novel intertextual work, generating observations and connections that appear to be unprecedented in the scholarly literature. This suggests that LLMs have the potential to uncover new insights and generate fresh ideas for biblical intertextual studies.


\subsection{Future Work}

An interesting facet of this paper is the inspiration from fuzzy neural networks which provided an important lens. However, an important limitation of any work that attempts to charactize the intertextual dependence weight between any pair of passages is that it will necessarily fail to determine relational direction since historical information is sparse. So, future work should simultaneously consider multiple nodes within the graph to identify the order of intertextual dependence which finds the most probable candidate textual chain.

Future work should also investigate the impact of contextual recall effects like the fan effect observed in some LLMs \cite{roberts2024large}. It may be that objects or people which appear frequently in the corpus in varying scenarios may be more highly impacted by LLM hallucinations leading to a greater frequency of intertextual false positives.

\section{Limitations}
There were some limitations in the LLM’s intertextual analysis which must be considered. The model struggled with long query passages, occasionally producing errors and non-sequiturs. It also failed to consider shared pretexts in some cases, asserting direct dependence between texts when a common source was more likely. Additionally, the intertextual candidates generated by the LLM sometimes included false positives or tenuous connections, requiring expert evaluation.

First, the model does not perform well with long queries. In the experiment with Acts 20 and Matthew 10, the model produced basic errors such as identifying words that did not exist and making non-sequitur judgments. The errors compounded as more prompts were submitted. 

Next, when presented with two similar texts for analysis, the model may not consider a shared pretext, even if one exists. This was the case with the analysis of Matthew 7 and 1 Peter 2:4-8. The model asserted direct depenence between the query and corpus without considering the shared pretexts in Psalms and Isaiah. 

Finally, the candidates generated by AI require the critical eye of an expert in the field. Of Hays’ seven criteria for intertextuality used for this paper, the LLM provided results based on volume, thematic coherence, and sensibility. The LLM did not evaluate availability, recurrence, historical plausibility, or the history of interpretation. The user must possess these competencies to properly evaluate generated candidates. 

\bibliography{anthology,custom}

\appendix


\end{document}